\pdfoutput=1

\documentclass[11pt,a4paper]{article}
\usepackage{arabtex}
\usepackage{utf8}
\usepackage[hyperref]{eacl2021}
\usepackage{times}
\usepackage{multirow}
\usepackage{latexsym}

\usepackage{microtype}

\aclfinalcopy 

\title{AraBERT and Farasa Segmentation Based Approach For Sarcasm and Sentiment Detection in Arabic Tweets}

\author{Anshul Wadhawan \\
  Flipkart Private Limited \\
  \texttt{anshul.wadhwan@flipkart.com} \\} 

\date{}

\begin{document}
\setcode{utf8}
\maketitle

\begin{abstract}

This paper presents our strategy to tackle the EACL WANLP-2021 Shared Task 2: Sarcasm and Sentiment Detection. One of the subtasks aims at developing a system that identifies whether a given Arabic tweet is sarcastic in nature or not, while the other aims to identify the sentiment of the Arabic tweet. We approach the task in two steps. The first step involves pre processing the provided ArSarcasm-v2 dataset by performing insertions, deletions and segmentation operations on various parts of the text. The second step involves experimenting with multiple variants of two transformer based models, AraELECTRA and AraBERT. Our final approach was ranked seventh and fourth in the Sarcasm and Sentiment Detection subtasks respectively.    

\end{abstract}

\section{Introduction}

During the last two decades, work on subjective language processing has been very common in literature. The work on sentiment analysis was a major theme that was pursued during this time. Sentiment Analysis is a process, according to \cite{liu}, in which we extract out and examine the emotional stance in a particular piece of text. With the introduction of user-driven networks such as websites for social media, the work on Sentiment Analysis flourished. Most of this work was based on English, while not much attention was gathered by the Arabic language. The work on Arabic Sentiment Analysis was initiated by \cite{abdul-mageed-etal-2011-subjectivity}, but as compared to English it still needs development. This can be due to the many complexities pertaining to the language, including the broad range of dialects \cite{10.1007/s10590-011-9087-8,darwish} and the complicated language morphology \cite{abdul-mageed-etal-2011-subjectivity}. Some of the prominent problems while tackling the task of Sentiment Analysis are domain dependency, negation handling, lack of sarcasm and knowledge \cite{HUSSEIN2018330}. Sarcasm is described as a form of verbal irony designed to convey disdain or ridicule \cite{10.1145/3124420}. There has been a lot of work on the identification of English sarcasm, including datasets, like the works of \cite{barbieri-etal-2014-modelling,ghosh-etal-2015-sarcastic,abercrombie-hovy-2016-putting,filatova-2012-irony,joshi-etal-2016-harnessing,barbieri} and detection systems like \cite{joshi-etal-2015-harnessing,10.1145/2684822.2685316,amir-etal-2016-modelling}. Work on Arabic sarcasm is, to our knowledge, restricted to the work of \cite{KAROUI2017161}, an irony detection task \cite{10.1145/3368567.3368585}, and sarcasm datasets by \cite{abbes-etal-2020-daict,abu-farha-magdy-2020-arabic}.

This paper puts forth the approach we applied to handle the WANLP-2021 Shared Task 2. The paper is ordered in the following manner: The problem statement, along with details of the ArSarcasm-v2 dataset are presented in Section 2. The methodology that we propose as our solution is described in Section 3. The experiments which were carried out, dataset statistics, system settings and results of the experiments are provided in Section 4. The paper ends with a brief section that talks about the conclusion and future directions of our research, in section 5.

\section{Task Definition}

The WANLP-2021 Shared Task 2 \cite{abufarha-etal-2021-arsarcasm-v2} is based on a text classification problem, based on identifying sentiment and sarcasm in Arabic tweets. The provided training and test datasets have a total of 12,548 and 3,000 tweets respectively.

The shared task is divided into two subtasks:

Subtask 1 (Sarcasm Detection): The aim is to identify whether a tweet is sarcastic or not. Given a tweet, the task is to return TRUE if there is sarcasm present in the tweet and FALSE otherwise. This is a binary classification problem. Precision/Recall/F-score/Accuracy are the evaluation metrics, where F-score of the sarcastic class is the official metric of evaluation.

Subtask 2 (Sentiment Analysis): The aim is to identify the sentiment of a tweet by assigning one of three labels (Positive, Negative, Neutral). Given a tweet, the task is to return POS if the tweet has a positive sentiment, NEG if the tweet has a negative sentiment or NEU if the tweet has a neutral sentiment. This is a multiclass classification problem. Precision/Recall/F-score/Accuracy are the evaluation metrics, where macro average of the F-score of the positive and negative classes (F-PN) is the official metric of evaluation.

\section{Methodology}

This section describes the process we employed to tackle the task. The process is divided into two steps: data preprocessing and transformer based models application. The first step involves processing the provided ArSarcasm-v2 dataset to convert it into a format better processed by the models. The second step involves experimenting with different models to decide which model performs the best for the ArSarcasm-v2 dataset. Details about these steps have been provided in the following sub-sections.

\subsection{Data Pre-Processing}

The data that is used for pre training the transformer based models is processed to create a better representation of the data. Thus, for the models to perform at their best ability, the data has to be processed in the same manner in the fine tuning process also. Raw data that is fetched from social media websites is diverse due to vast differences in expressions of opinions among users from different parts of the world. ArSarcasm-v2 dataset has these variations in different forms, evident from manual analysis. Social media users often make use of slang words, and non ascii characters such as emojis. URLs, user mentions and spelling errors are prominent in many posts on social media platforms. These attributes do not qualify as discerning features for classification tasks like Sarcasm and Sentiment Detection, and contribute to noise within the dataset. Therefore, we employ different pre processing techniques to clear this noise, so that the transformer based models only receive the relevant features. These techniques are as follows:

\begin{enumerate}
\item Remove HTML line breaks and markup, unwanted characters like emoticons, repeated characters ($>$ 2) and extra spaces.

\item Perform Farasa segmentation (for select models only) \cite{abdelali-etal-2016-farasa}.

\item Insert whitespace before and after all non Arabic digits or English Digits and Alphabet and the 2 brackets, and between words and numbers or numbers and words.

\item Replace all URLs with [ \< رابط > ], emails with [~\< بريد > ], mentions with [ \< مستخدم > ].

\end{enumerate}

\subsection{Transformer Based Models}

Deep learning methods have shown promising results in different machine learning domains such as Computer Vision \cite{krizhevsky} and Speech Recognition \cite{graves}. The traditional machine learning methods have been over taken by deep learning techniques in the recent past, because of their superior performance owing to architectures inspired by the human brain. On the lines of Natural Language Processing, most deep learning techniques have been making use of word vector representations \cite{yih2011,bengio,mikolov} mainly as a way of representing the textual inputs. These techniques are further being replaced by transformer based techniques \cite{vaswani2017attention} due to significant improvements on most NLP tasks like text classification \cite{chang2020taming}, which is the task at hand. Transformer based techniques have the ability to produce efficient embeddings as an output of the pre training process, which makes them proficient language models.

\begin{table*}[]
\begin{tabular}{|l|l|l|l|l|l|l|}
\hline
\multicolumn{1}{|c|}{\multirow{2}{*}{\textbf{Model}}} & \multicolumn{2}{c|}{\textbf{Size}}                                      & \multicolumn{1}{c|}{\multirow{2}{*}{\textbf{Pre-Segmentation}}} & \multicolumn{3}{c|}{\textbf{Dataset}}                                                                                  \\ \cline{2-3} \cline{5-7} 
\multicolumn{1}{|c|}{}                                & \multicolumn{1}{c|}{\textbf{MB}} & \multicolumn{1}{c|}{\textbf{Params}} & \multicolumn{1}{c|}{}                                           & \multicolumn{1}{c|}{\textbf{\#Sentences}} & \multicolumn{1}{c|}{\textbf{Size}} & \multicolumn{1}{c|}{\textbf{\#Words}} \\ \hline

AraBERTv0.1-base                                      & 543MB                            & 136M                                 & No                                                              & 77M                                       & 23GB                               & 2.7B                                  \\ \hline
AraBERTv1-base                                        & 543MB                            & 136M                                 & Yes                                                             & 77M                                       & 23GB                               & 2.7B                                  \\ \hline
AraBERTv0.2-base                                      & 543MB                            & 136M                                 & No                                                              & 200M                                      & 77GB                               & 8.6B                                  \\ \hline
AraBERTv0.2-large                                     & 1.38G                            & 371M                                 & No                                                              & 200M                                      & 77GB                               & 8.6B                                  \\ \hline
AraBERTv2-base                                        & 543MB                            & 136M                                 & Yes                                                             & 200M                                      & 77GB                               & 8.6B                                  \\ \hline
AraBERTv2-large                                       & 1.38G                            & 371M                                 & Yes                                                             & 200M                                      & 77GB                               & 8.6B                                  \\ \hline
\end{tabular}
\caption{Architectural Attributes of Models}
\end{table*}

\subsubsection{AraBERT} \cite{antoun2020arabert} Pre trained to handle Arabic text, AraBERT is a language model that is inspired from the Google's BERT architecture. Six variants of the same model are available for esxperimentation: AraBERTv0.2-base, AraBERTv1-base, AraBERTv0.1-base, AraBERTv2-large, AraBERTv0.2-large, and AraBERTv2-base. The architectural attributes of each of these models have been highlighted in Table 1.

\subsubsection{AraELECTRA} \cite{antoun2020araelectra} With reduced computations for pre training the transformers, ELECTRA is a method aimed towards the task of self-supervised language representation learning. ELECTRA models are inspired from the two primary components of Generative Adversarial Networks: generator and discriminator. They aim at distinguishing between real input tokens and the fake ones. These models have shown convincing state-of-the art results on Arabic QA data.

For the pretraining process of all new AraBERT and AraELECTRA models, the same data is used, which has a size of 77GB. It has in total 8,655,948,860 words or 82,232,988,358 characters or 200,095,961 lines, before Farasa segmentation. For an initial pre training dataset, several websites were crawled, which are: Assafir news articles, OSCAR unshuffled and filtered, The OSIAN Corpus, Arabic Wikipedia dump from 2020/09/01, and The 1.5B words Arabic Corpus. For newer models, a fresh dataset was developed for the pre training process, which did not include the data crawled from the above websites, but had unshuffled and properly filtered OSCAR corpus in addition to the dataset used in AraBERTv1. 

\section{Experiments}

We split the provided training dataset in a 90:10 ratio to create our training and validation splits. This is followed by experimenting with eight transformer based models, for which performance on the validation split is compared. The pre trained models are fine tuned on the training split and metrics for the validation split are calculated. The final test set predictions are made from the model that performs the best on the validation split, among the eight models. This section includes the dataset distribution, system settings, results and a brief analysis of the system, for both the subtasks: sarcasm and sentiment detection. 

\subsection{Dataset}

\begin{table}[]
\centering
\begin{tabular}{|l|l|l|l|}
\hline
\multicolumn{1}{|c|}{\multirow{2}{*}{\textbf{}}} & \multicolumn{2}{c|}{\textbf{Sarcasm}}                                    & \multicolumn{1}{c|}{\multirow{2}{*}{\textbf{Total}}} \\ \cline{2-3}
\multicolumn{1}{|c|}{}                           & \multicolumn{1}{c|}{\textbf{FALSE}} & \multicolumn{1}{c|}{\textbf{TRUE}} & \multicolumn{1}{c|}{}                                \\ \hline
Train                                            & 9356                                & 1937                               & 11293                                                \\ \hline
Dev                                              & 1024                                & 231                                & 1255                                                 \\ \hline
Total                                            & 10380                               & 2168                               & 12548                                                \\ \hline
\end{tabular}
\caption{Data Distribution w.r.t Sarcasm}
\end{table}

\begin{table}[]
\centering
\begin{tabular}{|l|l|l|l|l|}
\hline
\multicolumn{1}{|c|}{\multirow{2}{*}{\textbf{}}} & \multicolumn{3}{c|}{\textbf{Sentiment}}                                                                   & \multicolumn{1}{c|}{\multirow{2}{*}{\textbf{Total}}} \\ \cline{2-4}
\multicolumn{1}{|c|}{}                           & \multicolumn{1}{c|}{\textbf{NEG}} & \multicolumn{1}{c|}{\textbf{NEU}} & \multicolumn{1}{c|}{\textbf{POS}} & \multicolumn{1}{c|}{}                                \\ \hline
Train                                            & 4139                              & 5197                              & 1957                              & 11293                                                \\ \hline
Dev                                              & 482                               & 550                               & 223                               & 1255                                                 \\ \hline
Total                                            & 4621                              & 5747                              & 2180                              & 12548                                                \\ \hline
\end{tabular}
\caption{Data Distribution w.r.t Sentiment}
\end{table}

Tables 2 and 3 show the class wise distribution of the 90:10 training-validation splits created from the provided ArSarcasm-v2 dataset, for the tasks of sarcasm detection which has class labels TRUE and FALSE and sentiment detection which has class labels NEG, NEU and POS, respectively.

\subsection{System Settings}

\begin{table}[]
\centering
\begin{tabular}{|l|l|}
\hline
\textbf{Parameter}        & \textbf{Value}  \\ \hline
Epsilon (Adam optimizer)  & 1e-8            \\ \hline
Learning Rate             & 1e-5            \\ \hline
Batch Size (for base models)                & 40               \\ \hline
Batch Size (for large models)                & 4               \\ \hline
Maximum Sequence Length   & 256             \\ \hline
\#Epochs                  & 10               \\ \hline
\end{tabular}
\caption{Parameter Values}
\end{table}

\begin{table*}[]
\centering
\begin{tabular}{|l|l|l|l|l|l|l|l|l|}
\hline
\multicolumn{1}{|c|}{\multirow{2}{*}{\textbf{Model}}} & \multicolumn{4}{c|}{\textbf{Subtask 1}}                                                                                                & \multicolumn{4}{c|}{\textbf{Subtask 2}}                                                                                                \\ \cline{2-9} 
\multicolumn{1}{|c|}{}                                & \multicolumn{1}{c|}{\textbf{P}} & \multicolumn{1}{c|}{\textbf{R}} & \multicolumn{1}{c|}{\textbf{F1}} & \multicolumn{1}{c|}{\textbf{A}} & \multicolumn{1}{c|}{\textbf{P}} & \multicolumn{1}{c|}{\textbf{R}} & \multicolumn{1}{c|}{\textbf{F1}} & \multicolumn{1}{c|}{\textbf{A}} \\ \hline
AraBERTv0.1-base                                      & 84.15                           & 84.62                           & 84.36                            & 84.62                           & 72.38                           & 72.43                           & 72.40                            & 72.43                           \\ \hline
AraBERTv0.2-base                                      & 84.31                           & 85.42                           & 84.62                            & 85.42                           & 74.90                           & 74.98                           & 74.90                            & 74.98                           \\ \hline
AraBERTv0.2-large                                     & 84.57                           & 85.90                           & 84.63                            & 85.90                           & 76.25                           & 76.41                           & 76.20                            & 76.41                           \\ \hline
AraBERTv1-base                                        & 84.73                           & 85.02                           & 84.86                            & 85.02                           & 73.27                           & 73.55                           & 73.17                            & 73.55                           \\ \hline
AraBERTv2-base                                        & 85.39                           & 85.74                           & 85.55                            & 85.74                           & 75.93                           & 75.86                           & 75.89                            & 75.86                           \\ \hline
AraBERTv2-large                                       & 85.16                           & 86.37                           & 85.07                            & 86.37                           & 74.28                           & 74.50                           & 74.31                            & 74.50                           \\ \hline
AraELECTRA-base-generator                             & 82.72                           & 83.19                           & 82.94                            & 83.19                           & 72.45                           & 72.03                           & 71.99                            & 72.03                           \\ \hline
AraELECTRA-base-discriminator                         & 84.75                           & 85.34                           & 84.99                            & 85.34                           & 74.70                           & 74.82                           & 74.57                            & 74.82                           \\ \hline
\end{tabular}
\caption{Results on Validation Set}
\end{table*}

We make use of hugging-face\footnote{\url{https://huggingface.co/transformers/}} API to fetch the pre-trained AraBERT and AraELECTRA models. The API provides the six variants of AraBERT models by the names of bert-base-arabertv02, bert-base-arabert, bert-base-arabertv2, bert-base-arabertv01, bert-large-arabertv2, and bert-large-arabertv02, and the two variants of AraELECTRA models by the names of araelectra-base-discriminator and araelectra-base-generator. We fine tune these models on the training split with hyper parameter values specified in Table 4.

\subsection{Results and Analysis}

This section provides the detailed results obtained on the created validation and provided test sets.

\subsubsection{Validation Set Results}

The experimental results in terms of weighted Precision(P), weighted F1 scores(F1), weighted Recall(R) and Accuracy(A) on the created validation split have been depicted in Table 5, for both subtasks: Sarcasm Detection (Subtask 1) and Sentiment Detection (Subtask 2).

Our observations from Table 5 are as follows:
\begin{enumerate}

\item When comparing all the models, AraELECTRA generator model has the worst performance in terms of both F1 scores and Accuracy, for both the subtasks. This is possibly due to its forte of handling GAN related tasks rather than general classification tasks.

\item When comparing all AraBERT models, one of the large models seem to perform the best in terms of accuracy for both the subtasks. This is possibly due to superior architectures and heavier models. 

\item For Subtask 1, AraBERTv2-base has the highest weighted F1-score and for Subtask 2, AraBERTv0.2-large has the highest weighted F1-score. 

\end{enumerate}

\begin{table}[]
\centering
\begin{tabular}{|c|c|c|c|c|c|}
\hline
              & \textbf{C\textsubscript{E}} & \textbf{A} & \textbf{P} & \textbf{R} & \textbf{M-F1}\\ \hline
T\textsubscript{S}\textsuperscript{1} & 58.72  & 78.30 & 72.64 & 71.47 & 72.00\\ \hline
T\textsubscript{S}\textsuperscript{2} & 72.55 & 69.83  & 65.15 & 66.23 & 65.31\\ \hline

\end{tabular}
\caption{Official Results on Test Set}
\end{table}

\subsubsection{Test Set Results}

From the above observations, we select AraBERTv2-base for Subtask 1 and AraBERTv0.2-large for Subtask 2 as final models to formulate our officially submitted predictions. Table 6 presents the final test set results, with T\textsubscript{S}\textsuperscript{1} denoting Subtask 1, T\textsubscript{S}\textsuperscript{2} denoting Subtask 2, P denoting Precision, R denoting Recall, A denoting Accuracy, M-F1 denoting Macro-F1 score, and C\textsubscript{E} denoting the criteria of evaluation for the two subtasks (F-score of the sarcastic class for T\textsubscript{S}\textsuperscript{1}, and Macro averaged F score of positive and negative classes for T\textsubscript{S}\textsuperscript{2}). 

\section{Conclusion and Future Work}

In this paper, we present our strategy to approach the EACL WANLP-2021 Shared Task 2. We tackle the task in two steps. In the first step, the ArSarcasm-v2 dataset is pre-processed by altering different parts of text. This is followed by running experiments with multiple variants of two transformer based models pre-trained on Arabic text, AraELECTRA and AraBERT. The final submissions for the tasks of Sarcasm and Sentiment Detection are based on that variant of model which performs the best. Our approach fetches a private leaderboard rank 7 and 4 in the Sarcasm and Sentiment Detection tasks respectively. As future scope, we plan to explore other features which may be relevant for this task, and inculcate ensemble learning taking into consideration both word vector based and transformer based embeddings.

\bibliography{eacl2021}
\bibliographystyle{acl_natbib}

\end{document}